\documentclass{article} 
\usepackage{iclr2021_conference,times}

\usepackage{amsmath,amsfonts,bm}

\def\eqref#1{equation~\ref{#1}}

\def\1{\bm{1}}

\DeclareMathAlphabet{\mathsfit}{\encodingdefault}{\sfdefault}{m}{sl}
\SetMathAlphabet{\mathsfit}{bold}{\encodingdefault}{\sfdefault}{bx}{n}

\usepackage{hyperref}
\usepackage{url}

\usepackage[pdftex]{graphicx}
\title{CropGym: a reinforcement learning environment for crop management}

\author{Hiske Overweg\thanks{Corresponding author: hiske.overweg@wur.nl},~Herman~N.~C.~Berghuijs,~Ioannis~N.~Athanasiadis\\
Wageningen University and Research, The Netherlands\\}

\iclrfinalcopy 
\begin{document}

\hyphenation{Open-AI}

\maketitle

\begin{abstract}
Nitrogen fertilizers have a detrimental effect on the environment, which can be reduced by optimizing fertilizer management strategies. We implement an OpenAI Gym environment where a reinforcement learning agent can learn fertilization management policies using process-based crop growth models and identify policies with reduced environmental impact. In our environment, an agent trained with the Proximal Policy Optimization algorithm is more successful at reducing environmental impacts than the other baseline agents we present.
\end{abstract}

\section{Introduction}

Fertilizer use has vastly improved crop yields in the past century \citep{Frink1999}. Yet excessive use of fertilizers is damaging the environment in several ways \citep{Schlesinger2009,Sheriff2005,Yadav1997}. Leakage into waterways can lead to eutrification and lower oxygen levels up to an extent where marine life is no longer supported 
\citep{Diaz2008}. On top of that, excessive fertilization contributes to global warming due to high energy demand of fertilizer production and conversion by soil bacteria into nitrous oxide, a potent green house gas \citep{Erisman2011}.

By improved farming practises, leakage into the environment can be reduced. As an example, improvements can come from tailored side dress recommendations, where at the start of the season a small dose of fertilizer is applied which gets supplemented locally during the growth season \citep{VanEvert2012}. Alternatively, intercropping methods \citep{Cong2015} and other eco-innovations can reduce the need for synthetic fertilization \citep{Hasler2017}. 

Studying and validating farming strategies in practise is a challenging process, because of the long duration and labor intensity of field trials and the presence of many confounding factors. Process-based models, which simulate crop growth, can help to gain insights in this area. In this work, we create a reinforcement learning environment in which an agent can learn fertilization strategies using a process-based crop growth model. While reinforcement learning has had significant impact in other domains, we think its potential in agriculture has not been fully recognized yet. By providing an OpenAI Gym environment, we aim to encourage exploration of the application of reinforcement learning to sustainable agricultural management.

\section{Related work}

Deep RL is the subfield of machine learning which is concerned with training agents to take actions in an environment based on a reward signal \citep{Sutton2018b}. Deep RL has been successfully applied in numerous fields, ranging from computer games \citep{Silver2016,Mnih2013}, to the autonomous navigation of stratospheric balloons \citep{Bellemare2020}. Applications to agriculture have been proposed in \citet{Bu2019} and 
\citet{Binas2019}. RL agents have been implemented to control irrigation in a greenhouse setting \citep{Zhou2020} and to determine field sampling strategies using remote sensing \citep{Zhang2020a}. We are not aware of any works which study crop fertilization strategies or make use of process-based crop growth models.

OpenAI Gym provides a standardized collection of benchmark problems with a common interface \citep{Brockman2016}. It allows for comparison between algorithms and assessment of generalization performance.  It also provides an interface to convert a real-world problem into a custom Gym environment. 

Process-based crop growth models are designed for quantitative analysis of biophysical processes in crop growth and production. These models describe crop growth processes in terms of differential equations which estimate crop growth in daily timesteps, based on factors such as light interception, water and nutrient availability. Widely used frameworks are APSIM \citep{Keating2003,Holzworth2014} and the Python Crop Simulation Environment (PCSE), which contains models as LINTUL-3 \citep{Shibu2010} and WOFOST (WOrld FOod STudies, \citet{DeWit2019}). For a literature review of agricultural systems modelling, see \citet{Jones2017a}. Modern wheat crop models are discussed in \citet{Chenu2017}.

\section{Description of the Gym environment}

The CropGym environment\footnote{The source code is available at \url{https://github.com/BigDataWUR/crop-gym}} contains the following components:
\begin{itemize}
    \item The state space $S$, consisting of the current state of the crop (in terms of a multidimensional output of a process-based crop growth model) and an multidimensional weather observation
    \item The action space $A$, which consists of discrete doses of fertilizer to apply
    \item Transitions between states, governed by the deterministic process-based model and the weather sequence
    \item The reward $r$, which encourages large yield and limited fertilizer use.
    
\end{itemize}
\begin{figure}[h]
\begin{center}
\includegraphics[width=0.3 \textwidth]{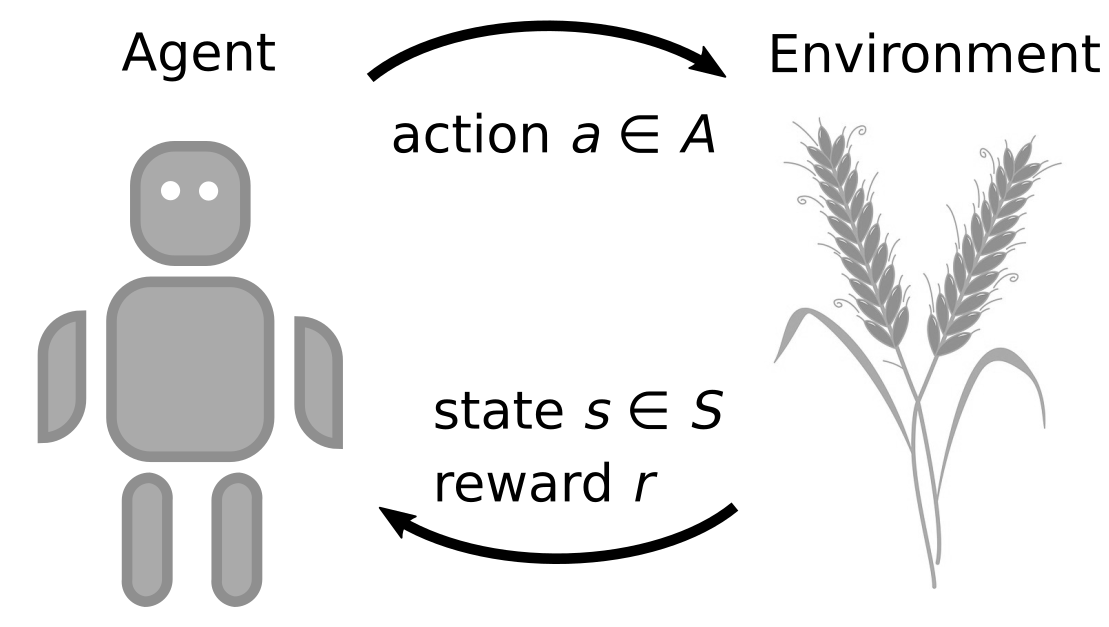}
\end{center}
\caption{Interaction between the agent and the CropGym environment}
\label{fig:rl}
\end{figure}
The information flow is shown schematically in Fig.~\ref{fig:rl}. The above framework can be applied for any process-based crop growth model or farm management simulation. In the next sections we detail our implementation choices.
\paragraph{State space}

A natural intervention interval for farmers is a week. This is why, using the default settings, the agent observes and intervenes in the CropGym environment on a weekly basis. The observations by the agent consist of the output variables of the process-based crop growth model LINTUL-3 (Light INtercepion and UTilization). Nitrogen limited crop growth in LINTUL-3 is implemented in the Python Crop Simution Environment (PCSE). The model parameters have been calibrated to simulate winter wheat \citep{Wiertsema2015}, which is why we chose to focus on this crop. The observed variables from the crop growth model are listed in Appendix \ref{sec:crop_vars}. Additionally the agent observes the weather of the past week as described by the default weather variables in PCSE (see Appendix \ref{sec:weather_vars}). For these weather variables we use 29 years of weather data from 9 locations in the Netherlands from the PowerNASA database.

\paragraph{Action space}

The agent's action space consists of a range of nitrogen fertilizer values to apply at each time step. From a farm management perspective, applying no fertilizer at all at a given point in time is an attractive choice, because there is no time investment. This is why we chose to discretize the action space. At each intervention, the agent has the following options for fertilizer application amounts:

\begin{equation}
   A =  \{20~k \frac{\mathrm{kg}}{\mathrm{ha}} ~|~k \in \{0, 1, 2, ..., 6\}\}
\end{equation}
In this way the agent can experiment with both doses which are smaller than and similar to those used in practise \citep{Wiertsema2015}.

\paragraph{Reward}
\label{sec:reward}

We reward the agent for a large grain yield (we do not take losses during harvest into account), which is referred to in LINTUL-3 as the mass of the storage organ ($m_{SO}$) of the crop, expressed in kilograms of dry matter per hectare. The achievable mass varies per season, because weather variables affect the yield. To account for this, we calculate both the mass $m_{SO}$ the agent achieves with its fertilization policy, and the mass $m^*_{SO}$ which would have been reached without fertilizer application. Our reward function contains the difference between these quantities. Finally, our goal is to discourage fertilizer application, which is why we penalize the agent for the weight of fertilizer applied ($m_{fert}$), expressed in the same units.  The reward at timestep $t$ is thus defined as:
\begin{equation*}
    r_t = m_{SO, t}- m_{SO, t-1} -(m^*_{SO, t} - m^*_{SO, t-1}) - \beta m_{fert, t}  
\end{equation*}
where the parameter $\beta$ determines the trade-off between large yield and reduced environmental impact. Setting $\beta \sim 1$ leads to optimization of the current economic cost for a farmer in the Netherlands, since wheat prices and fertilizer prices are similar \citep{AgrimatieGraan, AgrimatieMest}. When developing policies which have reduced environmental impact, we need to set $\beta > 1$.

\section{Baseline agents}
\paragraph{Implemented agents}
To determine whether reinforcement learning agents can indeed identify efficient fertilization policies, we compare three agents:
\begin{itemize}
    \item The standard practise agent adds applies nitrogen fertilizer on three different dates during the season, as is done in practise in the Netherlands. For details, see Appendix \ref{sec:agent}.
    \item  The reactive agent adds a fixed amount of fertilizer to the field whenever the nitrogen in the soil is depleted.
    \item The PPO agent is trained with the Proximal Policy Optimization algorithm, a widely used policy gradient algorithm \citep{Schulman2017}, as implemented in the Stable Baselines3 package \citep{stable-baselines3}.
    The policy and value function networks are both fully connected networks consisting of 2 layers with 64 nodes per layer. \end{itemize}
    For more details on agent training and hyperparameters, see Appendix \ref{sec:agent}.

\paragraph{Agent performance}
\begin{figure}[h]
\begin{center}
\includegraphics[width=\textwidth]{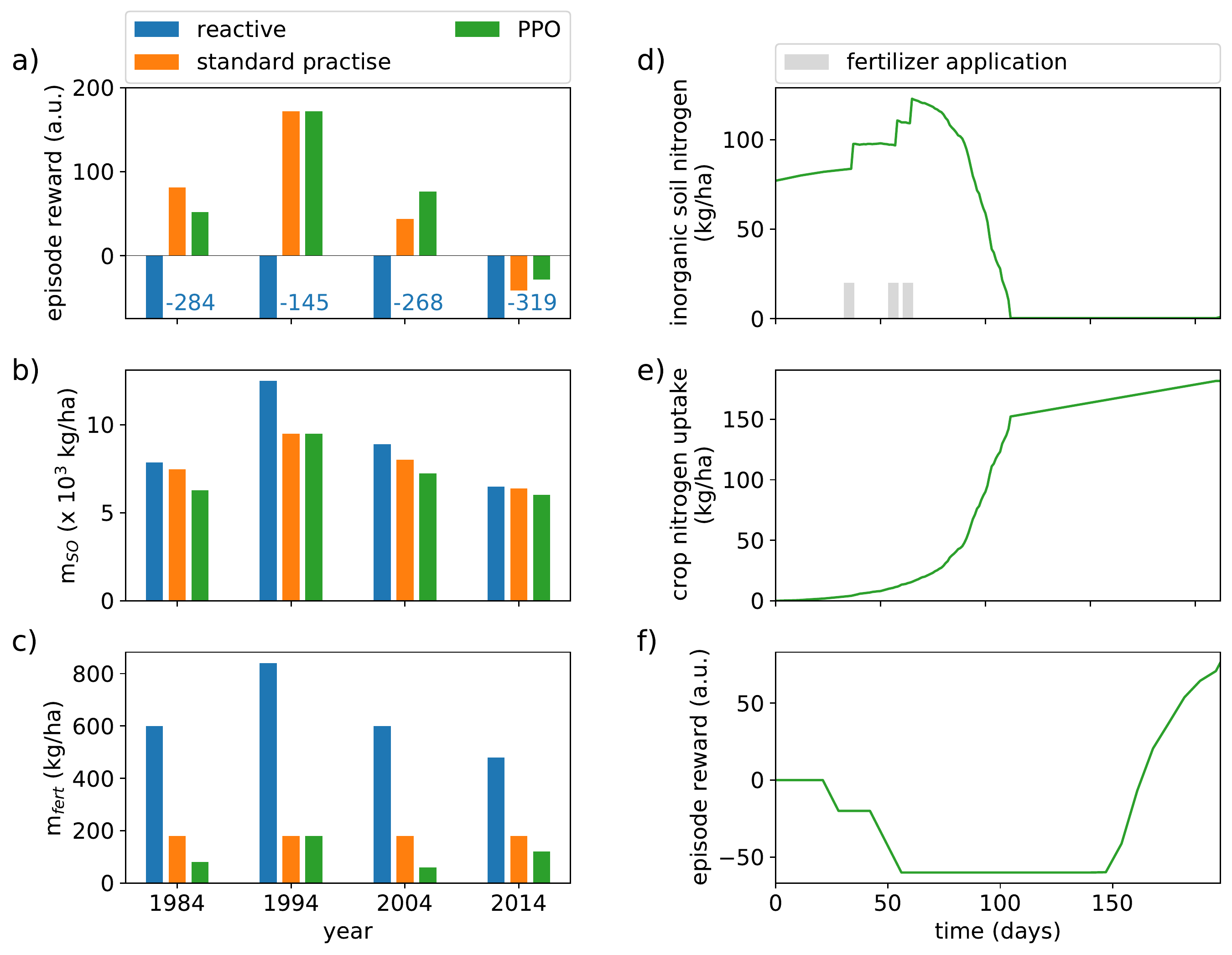}
\end{center}
\caption{Agent performance. (a) Reward obtained by various agents. (b) The weight of the storage organ at the end of the season. (c) The amount of fertilizer applied by the agents. (d) Nitrogen available in the soil as a function of time in the season starting in 2004. Gray bars indicate fertilization events by the PPO agent.  (e) Wheat nitrogen uptake. (f) Reward of the PPO agent.}
\label{fig:reward}
\end{figure}

We evaluate the reward the agents obtain on the seasons starting in 1984, 1994, 2004 and 2014 with weather data from 52$^o$N, 5.5$^o$E. These years were excluded from the training set of the PPO agent. For the reward function we set $\beta=10$. As can be seen in Fig.~\ref{fig:reward}a, the PPO agent gets a higher reward than the standard practise agent and the reactive agent in 1994, 2004 and 2014. In Fig.~\ref{fig:reward}b,c we show the total yield obtained and the amounts of fertilizer applied by the agents respectively. The PPO agent has a somewhat smaller yield than the other agents in 1984 and 2004, because of the reduced fertilization. Figure~\ref{fig:reward}d shows the soil nitrogen level and the fertilization events proposed by the PPO agent in 2014. The nitrogen uptake by the wheat and the reward are shown in Fig.~\ref{fig:reward}e,f respectively. Similar data for other years from the test set is shown in Appendix~\ref{sec:results}.

\section{Conclusion}

We introduce a new Gym environment to study fertilization strategies for crops and discuss the considerations we made during the design process of the CropGym. We implement a reactive agent, standard practise and an agent trained with PPO, which serve as a baseline for further investigations. The PPO agent was able to learn strategies with a higher reward than the other agents for three out of four years in the test set, which demonstrates that reinforcement learning agents are able to identify policies with reduced environmental impact.

\section{Future work}

\paragraph{Other farm management practices}

After further calibration of the more detailed process-based model WOFOST, we intend to train an agent to learn policies for more management options, as  phosphorus and potassium fertilization. Learning irrigation policies and including weather forecast data are other extensions we consider. The potential of reinforcement learning agents can be further exploited at a large scale when using whole farm simulations, which take interactions between fields and livestock farming into account \citep{Holzworth2014, Rodriguez2014}.

\paragraph{Environmental impact and climate change}
When coupling process-based crop growth models to more detailed soil models, the interaction between fertilization, rainfall and irrigation can be taken into account \citep{Engel1993, Groenendijk2016}. Just as process-based models, the CropGym can be used to study the effect of a changing climate on crop growth, to assess adaptations of farm management strategies  \citep{Rodriguez2014}.

\paragraph{Reality gap}
To get closer to the practical situation in the field, the environment can be modified to a partially observable Markov Decision Process, in which the agent only observes variables which can be realistically observed on a weekly basis. More accurate digital representations of crop growth processes would also help to get closer to reality. These can be provided by tailored calibration of process-based crop growth models, inclusion of other experimental data sources within the models \citep{Holzworth2014} or digital farm twins \citep{Jans-Singh2020, Pylianidis2021}.

\subsubsection*{Acknowledgments}
We thank Diego Marcos Gonzalez, Sjoukje Osinga, Pytrik Reidsma, Allard de Wit, Rob Knapen, Frits van Evert and Bernardo Maestrini for fruitful discussions. This work has been supported by the Wageningen University and Research Investment Programme “Digital Twins”.

\bibliography{bibliography_v2}

\begin{thebibliography}{34}
\providecommand{\natexlab}[1]{#1}
\providecommand{\url}[1]{\texttt{#1}}
\expandafter\ifx\csname urlstyle\endcsname\relax
  \providecommand{\doi}[1]{doi: #1}\else
  \providecommand{\doi}{doi: \begingroup \urlstyle{rm}\Url}\fi

\bibitem[agrimatie.nl({\natexlab{a}})]{AgrimatieGraan}
agrimatie.nl.
\newblock Price development of seeds and grains
  \url{https://www.agrimatie.nl/SectorResultaat.aspx?subpubID=2232&sectorID=2233&themaID=2263}.
\newblock {\natexlab{a}}.
\newblock Accessed: 2021-03-09.

\bibitem[agrimatie.nl({\natexlab{b}})]{AgrimatieMest}
agrimatie.nl.
\newblock Price development of fertilizer
  \url{https://www.agrimatie.nl/ThemaResultaat.aspx?subpubID=2289&themaID=2263}.
\newblock {\natexlab{b}}.
\newblock Accessed: 2021-03-09.

\bibitem[Bellemare et~al.(2020)Bellemare, Candido, Castro, Gong, Machado,
  Moitra, Ponda, and Wang]{Bellemare2020}
Marc~G. Bellemare, Salvatore Candido, Pablo~Samuel Castro, Jun Gong, Marlos~C.
  Machado, Subhodeep Moitra, Sameera~S. Ponda, and Ziyu Wang.
\newblock {Autonomous navigation of stratospheric balloons using reinforcement
  learning}.
\newblock \emph{Nature}, 588\penalty0 (7836):\penalty0 77--82, 2020.

\bibitem[Binas et~al.(2019)Binas, Luginbuehl, and Bengio]{Binas2019}
Jonathan Binas, Leonie Luginbuehl, and Yoshua Bengio.
\newblock {Reinforcement Learning for Sustainable Agriculture}.
\newblock \emph{CCAI workshop at the 36 th International Conference on Machine
  Learning}, 2019.

\bibitem[Brockman et~al.(2016)Brockman, Cheung, Pettersson, Schneider,
  Schulman, Tang, and Zaremba]{Brockman2016}
Greg Brockman, Vicki Cheung, Ludwig Pettersson, Jonas Schneider, John Schulman,
  Jie Tang, and Wojciech Zaremba.
\newblock {OpenAI Gym}.
\newblock pp.\  1--4, 2016.
\newblock URL \url{http://arxiv.org/abs/1606.01540}.

\bibitem[Bu \& Wang(2019)Bu and Wang]{Bu2019}
Fanyu Bu and Xin Wang.
\newblock {A smart agriculture IoT system based on deep reinforcement
  learning}.
\newblock \emph{Future Generation Computer Systems}, 99:\penalty0 500--507,
  2019.

\bibitem[Chenu et~al.(2017)Chenu, Porter, Martre, Basso, Chapman, Ewert, Bindi,
  and Asseng]{Chenu2017}
Karine Chenu, John~Roy Porter, Pierre Martre, Bruno Basso, Scott~Cameron
  Chapman, Frank Ewert, Marco Bindi, and Senthold Asseng.
\newblock {Contribution of Crop Models to Adaptation in Wheat}.
\newblock \emph{Trends in Plant Science}, 22\penalty0 (6):\penalty0 472--490,
  2017.

\bibitem[Cong et~al.(2015)Cong, Hoffland, Li, Six, Sun, Bao, Zhang, and {Van
  Der Werf}]{Cong2015}
Wen~Feng Cong, Ellis Hoffland, Long Li, Johan Six, Jian~Hao Sun, Xing~Guo Bao,
  Fu~Suo Zhang, and Wopke {Van Der Werf}.
\newblock {Intercropping enhances soil carbon and nitrogen}.
\newblock \emph{Global Change Biology}, 21\penalty0 (4):\penalty0 1715--1726,
  2015.

\bibitem[de~Wit et~al.(2019)de~Wit, Boogaard, Fumagalli, Janssen, Knapen, van
  Kraalingen, Supit, van~der Wijngaart, and van Diepen]{DeWit2019}
Allard de~Wit, Hendrik Boogaard, Davide Fumagalli, Sander Janssen, Rob Knapen,
  Daniel van Kraalingen, Iwan Supit, Raymond van~der Wijngaart, and Kees van
  Diepen.
\newblock {25 years of the WOFOST cropping systems model}.
\newblock \emph{Agricultural Systems}, 168\penalty0 (June 2018):\penalty0
  154--167, 2019.

\bibitem[Diaz \& Rosenberg(2008)Diaz and Rosenberg]{Diaz2008}
Robert~J. Diaz and Rutger Rosenberg.
\newblock {Spreading dead zones and consequences for marine ecosystems}.
\newblock \emph{Science}, 321\penalty0 (5891):\penalty0 926--929, 2008.

\bibitem[Engel \& Priesack(1993)Engel and Priesack]{Engel1993}
Thomas Engel and Eckart Priesack.
\newblock Expert-n-a building block system of nitrogen models as resource for
  advice, research, water management and policy.
\newblock In \emph{Integrated soil and sediment research: A basis for proper
  protection}, pp.\  503--507. Springer, 1993.

\bibitem[Erisman et~al.(2011)Erisman, Galloway, Seitzinger, Bleeker, and
  Butterbach-Bahl]{Erisman2011}
Jan~Willem Erisman, Jim Galloway, Sybil Seitzinger, Albert Bleeker, and Klaus
  Butterbach-Bahl.
\newblock {Reactive nitrogen in the environment and its effect on climate
  change}.
\newblock \emph{Current Opinion in Environmental Sustainability}, 3\penalty0
  (5):\penalty0 281--290, 2011.

\bibitem[Frink et~al.(1999)Frink, Waggoner, and Ausubel]{Frink1999}
Charles~R. Frink, Paul~E. Waggoner, and Jesse~H. Ausubel.
\newblock {Nitrogen fertilizer: Retrospect and prospect}.
\newblock \emph{Proceedings of the National Academy of Sciences of the United
  States of America}, 96\penalty0 (4):\penalty0 1175--1180, 1999.

\bibitem[Groenendijk et~al.(2016)Groenendijk, Boogaard, Heinen, Kroes, Supit,
  and de~Wit]{Groenendijk2016}
Piet Groenendijk, Hendrik Boogaard, Marius Heinen, JG~Kroes, Iwan Supit, and
  Allard de~Wit.
\newblock Simulation nitrogen-limited crop growth with swap/wofost: process
  descriptions and user manual.
\newblock Technical report, Wageningen Environmental Research, 2016.

\bibitem[Hasler(2017)]{Hasler2017}
Kathrin Hasler.
\newblock \emph{Environmental impact of mineral fertilizers: possible
  improvements through the adoption of eco-innovations}.
\newblock PhD thesis, Wageningen University, 2017.

\bibitem[Holzworth(2014)]{Holzworth2014}
Dean P. et~al. Holzworth.
\newblock {APSIM - Evolution towards a new generation of agricultural systems
  simulation}.
\newblock \emph{Environmental Modelling and Software}, 62:\penalty0 327--350,
  2014.

\bibitem[Jans-Singh et~al.(2020)Jans-Singh, Leeming, Choudhary, and
  Girolami]{Jans-Singh2020}
Melanie Jans-Singh, Kathryn Leeming, Ruchi Choudhary, and Mark Girolami.
\newblock {Digital twin of an urban-integrated hydroponic farm}.
\newblock \emph{Data-Centric Engineering}, 1, 2020.

\bibitem[Jones(2017)]{Jones2017a}
James W. et~al. Jones.
\newblock {Brief history of agricultural systems modeling}.
\newblock \emph{Agricultural Systems}, 155:\penalty0 240--254, 2017.

\bibitem[Keating(2003)]{Keating2003}
B.~A. et~al. Keating.
\newblock {An overview of APSIM, a model designed for farming systems
  simulation}.
\newblock \emph{European Journal of Agronomy}, 18\penalty0 (3-4):\penalty0
  267--288, 2003.
\newblock \doi{10.1016/S1161-0301(02)00108-9}.

\bibitem[Mnih et~al.(2013)Mnih, Kavukcuoglu, Silver, Graves, Antonoglou,
  Wierstra, and Riedmiller]{Mnih2013}
Volodymyr Mnih, Koray Kavukcuoglu, David Silver, Alex Graves, Ioannis
  Antonoglou, Daan Wierstra, and Martin Riedmiller.
\newblock {Playing Atari with Deep Reinforcement Learning}.
\newblock pp.\  1--9, 2013.

\bibitem[Pylianidis et~al.(2021)Pylianidis, Osinga, and
  Athanasiadis]{Pylianidis2021}
Christos Pylianidis, Sjoukje Osinga, and Ioannis~N Athanasiadis.
\newblock {Introducing digital twins to agriculture}.
\newblock \emph{Computers and Electronics in Agriculture}, 184\penalty0
  (December 2020):\penalty0 105942, 2021.

\bibitem[Raffin et~al.(2019)Raffin, Hill, Ernestus, Gleave, Kanervisto, and
  Dormann]{stable-baselines3}
Antonin Raffin, Ashley Hill, Maximilian Ernestus, Adam Gleave, Anssi
  Kanervisto, and Noah Dormann.
\newblock Stable baselines3.
\newblock \url{https://github.com/DLR-RM/stable-baselines3}, 2019.

\bibitem[Rodriguez et~al.(2014)Rodriguez, Cox, DeVoil, and
  Power]{Rodriguez2014}
Daniel Rodriguez, Howard Cox, Peter DeVoil, and Brendan Power.
\newblock {A participatory whole farm modelling approach to understand impacts
  and increase preparedness to climate change in Australia}.
\newblock \emph{Agricultural Systems}, 126:\penalty0 50--61, 2014.

\bibitem[Schlesinger(2009)]{Schlesinger2009}
William~H. Schlesinger.
\newblock {On the fate of anthropogenic nitrogen}.
\newblock \emph{Proceedings of the National Academy of Sciences of the United
  States of America}, 106\penalty0 (1):\penalty0 203--208, 2009.

\bibitem[Schulman et~al.(2017)Schulman, Wolski, Dhariwal, Radford, and
  Klimov]{Schulman2017}
John Schulman, Filip Wolski, Prafulla Dhariwal, Alec Radford, and Oleg Klimov.
\newblock {Proximal policy optimization algorithms}.
\newblock \emph{arXiv}, pp.\  1--12, 2017.

\bibitem[Sheriff(2005)]{Sheriff2005}
Glenn Sheriff.
\newblock {Efficient waste? Why farmers over-apply nutrients and the
  implications for policy design}.
\newblock \emph{Review of Agricultural Economics}, 27\penalty0 (4):\penalty0
  542--557, 2005.

\bibitem[Shibu et~al.(2010)Shibu, Leffelaar, van Keulen, and
  Aggarwal]{Shibu2010}
M.~E. Shibu, P.~A. Leffelaar, H.~van Keulen, and P.~K. Aggarwal.
\newblock {LINTUL3, a simulation model for nitrogen-limited situations:
  Application to rice}.
\newblock \emph{European Journal of Agronomy}, 32\penalty0 (4):\penalty0
  255--271, 2010.

\bibitem[Silver et~al.(2016)Silver, Schrittwieser, Simonyan, Nature, and
  2017]{Silver2016}
D~Silver, J~Schrittwieser, K~Simonyan, I~Antonoglou Nature, and Undefined 2017.
\newblock {Mastering the game of Go without human knowledge}.
\newblock \emph{Nature}, 550\penalty0 (7676):\penalty0 354, 2016.

\bibitem[Sutton(2018)]{Sutton2018b}
Richard~S. Sutton.
\newblock \emph{{Reinforcement Learning an introduction}}.
\newblock 2018.
\newblock ISBN 9780262039246.

\bibitem[van Evert et~al.(2012)van Evert, Booij, Jukema, ten Berge, Uenk,
  Meurs, van Geel, Wijnholds, and Slabbekoorn]{VanEvert2012}
Frits~K. van Evert, Remmie Booij, Jan~Nammen Jukema, Hein~F.M. ten Berge, Dik
  Uenk, E.~J.J.Bert Meurs, Willem~C.A. van Geel, Klaas~H. Wijnholds, and
  J.~J.Hanja Slabbekoorn.
\newblock {Using crop reflectance to determine sidedress N rate in potato saves
  N and maintains yield}.
\newblock \emph{European Journal of Agronomy}, 43:\penalty0 58--67, 2012.

\bibitem[Wiertsema(2015)]{Wiertsema2015}
Wiert Wiertsema.
\newblock {Obtaining winter wheat parameters for LINTUL from a field
  experiment}.
\newblock 2015.

\bibitem[Yadav et~al.(1997)Yadav, Peterson, and Easter]{Yadav1997}
Satya~N. Yadav, Willis Peterson, and K.~William Easter.
\newblock {Do farmers overuse nitrogen fertilizer to the detriment of the
  environment?}
\newblock \emph{Environmental and Resource Economics}, 9\penalty0 (3):\penalty0
  323--340, 1997.

\bibitem[Zhang et~al.(2020)Zhang, Boubin, Stewart, and Khanal]{Zhang2020a}
Zichen Zhang, Jayson Boubin, Christopher Stewart, and Sami Khanal.
\newblock {Whole-field reinforcement learning: A fully autonomous aerial
  scouting method for precision agriculture}.
\newblock \emph{Sensors}, 20\penalty0 (22):\penalty0 1--16, 2020.

\bibitem[Zhou(2020)]{Zhou2020}
Ni~Zhou.
\newblock {Intelligent control of agricultural irrigation based on
  reinforcement learning}.
\newblock \emph{Journal of Physics: Conference Series}, 1601\penalty0 (5),
  2020.

\end{thebibliography}
\bibliographystyle{iclr2021_conference}

\appendix
\renewcommand\thefigure{\thesection.\arabic{figure}}
\setcounter{figure}{0} 
\section{Appendix}

\subsection{Crop growth variables}
\begin{center}
\begin{tabular}{l| l| l }
 \textbf{Symbol} & \textbf{Meaning} & \textbf{Unit} \\ 
 \hline
DVS & development stage& - \\  
LAI & Leave area index & - \\
TGROWTH & Total biomass growth & g/m$^2$\\
NUPTT & Total N uptake & g N/m$^2$\\
TRAN & Crop transpiration rate & m$^3$ H$_2$O/m$^2$\\
TIRRIG & Water applied by irrigation & m$^3$ H$_2$O/m$^2$\\
TNSOIL & Soil nitrogen available & g N/m$^2$\\
TRAIN & Total rainfall & m$^3$ H$_2$O/m$^2$\\
TRANRF & Transpiration reduction factor & mm/day\\
TRUNOF & Soil runoff & m$^3$ H$_2$O/m$^2$\\
TAGBM & Total above ground dry weight & g/m$^2$\\
TTRAN & Water removed by transpiration & m$^3$ H$_2$O/m$^2$\\
WC & Soil moisture content & m$^3$ H$_2$O/m$^2$\\
WLVD & Dry weight of dead leaves & g/m$^2$\\
WRT & Dry weight of roots & g/m$^2$\\
WLVG & Weight green leaves & g/m$^2$\\
WSO & Dry weight of storage organ & g/m$^2$\\
WST & Dry weight of stems & g/m$^2$\\

\end{tabular}
\end{center}

\label{sec:crop_vars}
\subsection{Weather variables}
\label{sec:weather_vars}
\begin{center}
\begin{tabular}{l| l| l }
 \textbf{Symbol} & \textbf{Meaning} & \textbf{Unit} \\ 
 \hline
TMAX & Daily maximum temperature & $^o$C\\
TMIN & Daily minimum temperature & $^o$C\\
VAP & Mean daily vapour pressure & hPa\\
RAIN & Precipitation & cm/day \\
IRRAD & Daily global radiation & J/m$^2$/day

\end{tabular}
\end{center}

\subsection{Agent hyperparameters}
\label{sec:agent}

\paragraph{Standard practise agent}
The standard practise agent applied nitrogen on the following days, in correspondence with \cite{Wiertsema2015}:
\begin{center}
\begin{tabular}{l| l}
 \textbf{Date} & \textbf{Amount of fertilizer} \\ 
 \hline
3rd of March & 60 kg/ha\\
31st of March & 60 kg/ha\\
5th of May & 60 kg/ha

\end{tabular}
\end{center}

\paragraph{Reactive agent}
The reactive agent adds 120 kg/ha of nitrogen fertilizer whenever the soil nitrogen drops below 5 kg/ha.

\paragraph{PPO agent}
The environment was normalized with the VecNormalize environment wrapper, with a discount factor of $\gamma$=1, a normalized reward and observation clipping set to 10. The PPO agent trained for 500.000 timesteps using default hyperparameters.

\clearpage
\subsection{Agent Results}
\label{sec:results}

\begin{figure}[h]
\begin{center}
\includegraphics[width=\textwidth]{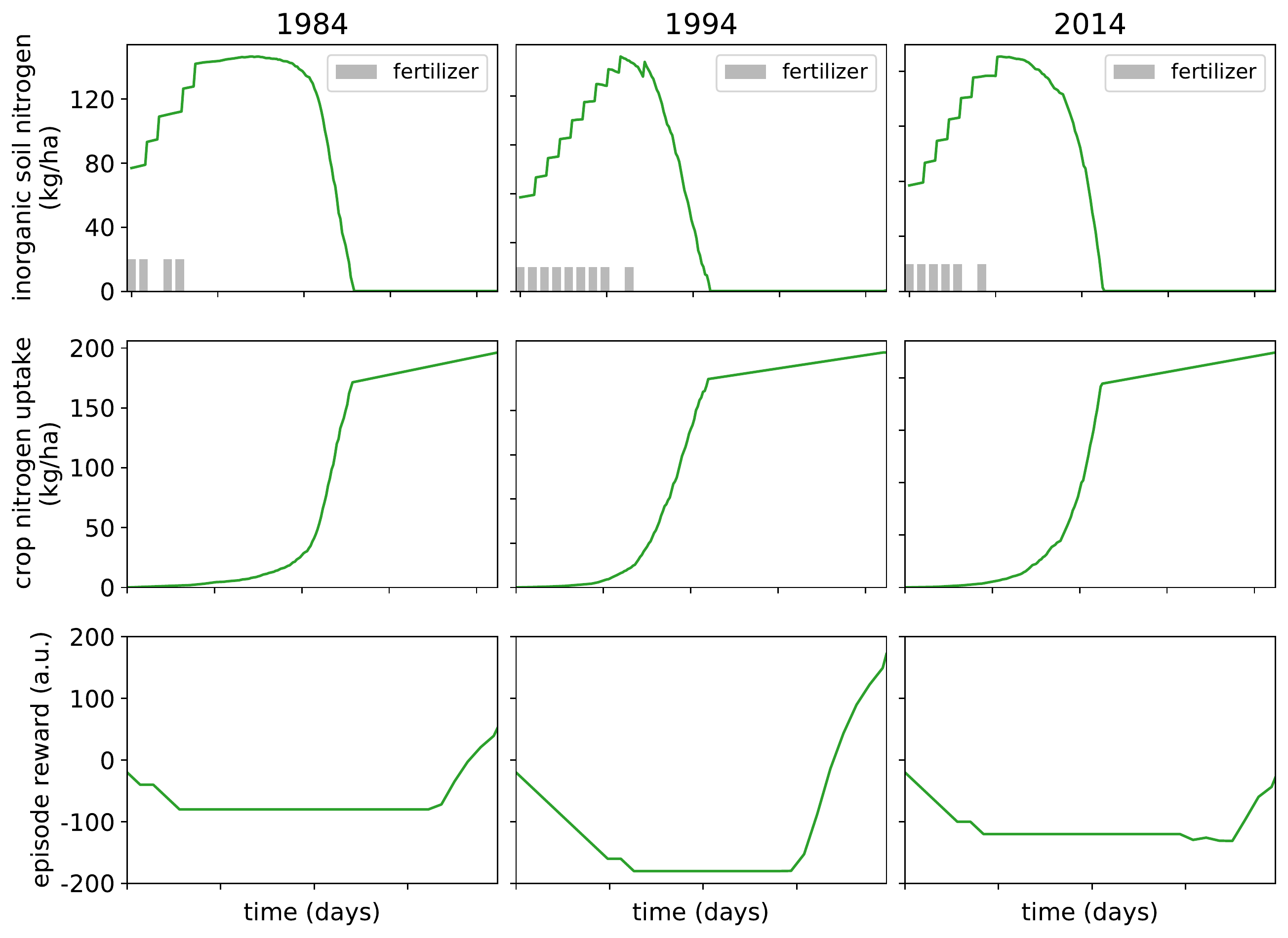}
\end{center}
\caption{Performance of the PPO agent in 1984, 1994 and 2014}
\label{fig:reward_other_years}
\end{figure}

\end{document}